\documentclass{ecai} 

\usepackage{latexsym}
\usepackage{amssymb}
\usepackage{amsmath}
\usepackage{amsthm}
\usepackage{booktabs}
\usepackage{enumitem}
\usepackage{graphicx}
\usepackage{color}

\newcommand{\BibTeX}{B\kern-.05em{\sc i\kern-.025em b}\kern-.08em\TeX}

\begin{document}


\begin{frontmatter}


\paperid{123} 


\title{Proactive Defense: Compound AI for Detecting Persuasion Attacks and Measuring Inoculation Effectiveness}


\author{\fnms{Svitlana}~\snm{Volkova}}
\author{\fnms{Will}~\snm{Dupree}}
\author{\fnms{Hsien-Te}~\snm{Kao}}
\author{\fnms{Peter}~\snm{Bautista}}
\author{\fnms{Gabe}~\snm{Ganberg}}
\author{\fnms{Jeff}~\snm{Beaubien}}
\author{\fnms{Laura}~\snm{Cassani}}


\address[]{Aptima, Inc.}


\begin{abstract}
 This paper introduces BRIES (Building Resilient Information Ecosystems), a novel compound AI architecture designed to detect and measure the effectiveness of persuasion attacks across information environments. Grounded in McGuire's (1964) inoculation theory and dual-process theories of persuasion~\cite{PettyCacioppo1986,Chaiken1980}, our framework extends these psychological approaches to complex digital environments through proactive defense strategies. As language models increasingly generate persuasive content at scale, developing proactive defense strategies becomes critical. We present a system with specialized agents: a Twister that generates adversarial content employing targeted persuasion tactics, a Detector that identifies attack types with configurable parameters, a Defender that creates resilient content through content inoculation, and an Assessor that employs causal inference to evaluate inoculation effectiveness. Experimenting with the SemEval 2023 Task 3 taxonomy across the synthetic persuasion dataset, we demonstrate significant variations in detection performance across language agents. Our comparative analysis reveals significant performance disparities with GPT-4 achieving superior detection accuracy on complex persuasion techniques like Appeal to Fear and Whataboutism, while open-source models like Llama3 and Mistral demonstrated notable weaknesses in identifying subtle rhetorical manipulations such as Equivocation and Red Herring , suggesting that different architectures encode and process persuasive language patterns in fundamentally different ways. We show that prompt engineering dramatically affects detection efficacy, with temperature settings and confidence scoring producing model-specific variations—Gemma and GPT-4 perform optimally at lower temperatures while Llama3 and Mistral show improved capabilities at higher temperatures. Our causal analysis provides novel insights into socio-emotional-cognitive signatures of persuasion attacks, revealing that different attack types target specific cognitive dimensions, with authority-based attacks showing distinct causal pathways compared to emotional appeals. This research advances generative AI safety and cognitive security by quantifying LLM-specific vulnerabilities to persuasion attacks and delivers a framework for enhancing human cognitive resilience through structured interventions before exposure to harmful content.
\end{abstract}
\end{frontmatter}


\section{Introduction}
The Building Resilient Information Ecosystems framework presents an innovative approach to addressing the growing challenge of targeted information campaigns that manipulate content and undermine trust in media and communications \cite{GAO2022, Deppe2024,NATOACT2023,Wardle2018,NATOFuture2024}. In today's complex information landscape, adversaries exploit generative AI to create campaigns targeting individuals, organizations, and agencies, particularly in domains where public perception is critical \cite{Goldstein2023}.

Traditional approaches to combating information manipulation have primarily focused on reactive measures, such as removing false content or accounts, or on general educational initiatives to increase awareness \cite{BakColeman2022}. However, these strategies fail to provide scalable, proactive solutions against emerging types of attacks. BRIES offers an alternative by leveraging recent developments in large language models (LLMs) \cite{Bommasani2021}, AI agents \cite{Park2023} and causal analysis \cite{Volkova2021b} to develop generative inoculation capabilities leveraging inoculation theory~\cite{Lewandowsky2021,VanderLinden2017, McGuire1964}, that build resilience against information operations.

 \begin{figure}[t!]
 \centering
  \includegraphics[width=0.5\textwidth]{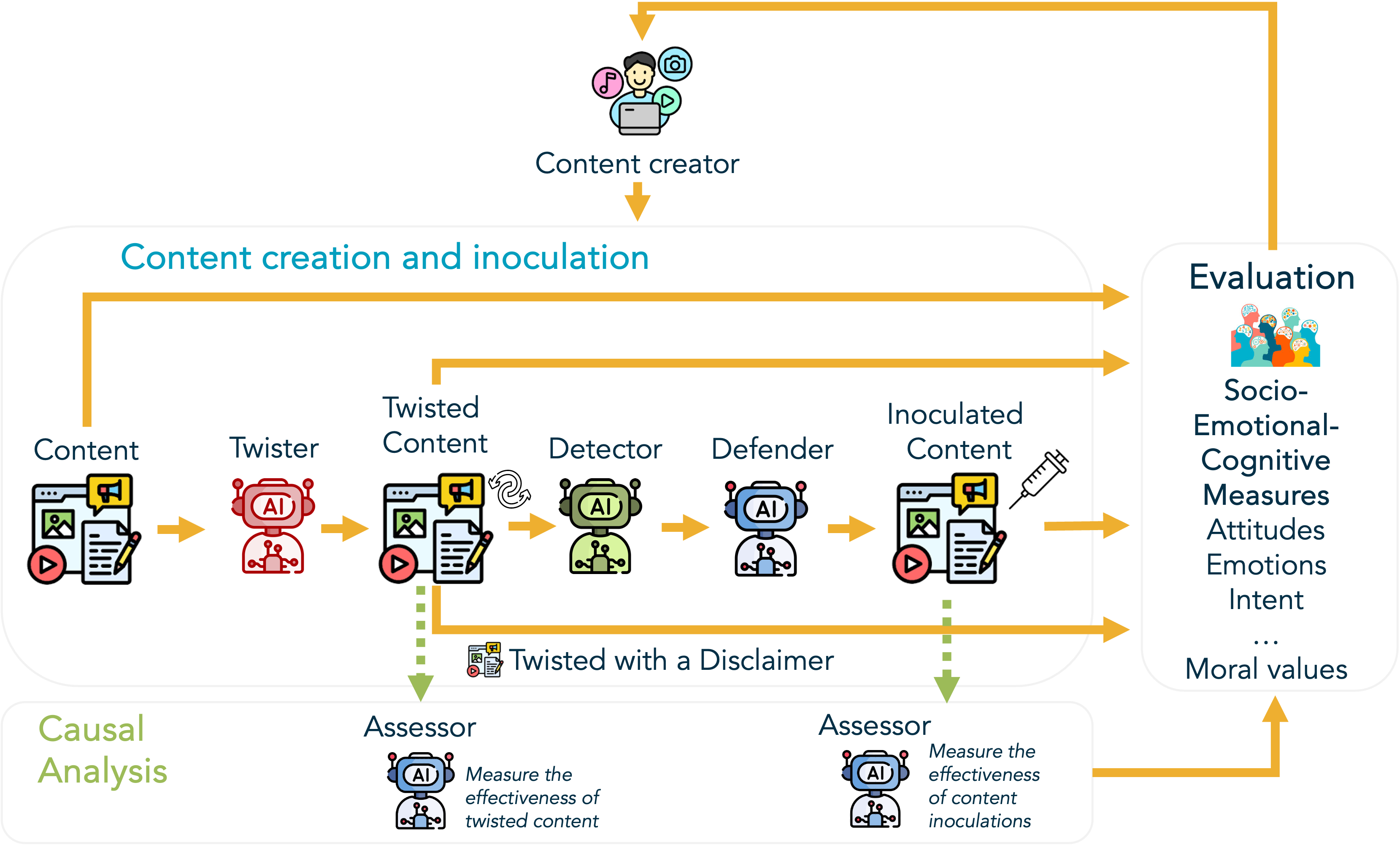}
 \caption{Building Resilient Information Ecosystems framework for content inoculations to enable rapid experimentation and measurement of effectiveness for proactive defense.}
 \label{fig:bries}
 \end{figure}

At the core of BRIES is a compound AI framework~\cite{compound-ai-blog} that begins with original content from creators and processes it through specialized language models, agents and tools (Figure~\ref{fig:bries}). The Twister Agent generates manipulated or "twisted" content using various persuasion techniques (note, in this paper we have used the existing synthetic dataset\footnote{\url{https://github.com/apanasyu/UNCOVER_SPIE}}; thus the Twister agent was not included into the experiments); the Detector Agent identifies attack types and vulnerable points in the content; the Defender Agent creates inoculated content resistant to manipulation; and the Assessor tools measure the effectiveness of persuasion attacks, detector models and content inoculations throughout the process.

For evaluation, the system employs causal analysis tools to validate content protection and uses a suite of socio-emotional-cognitive measures to evaluate effectiveness, including perspective analysis, toxicity detection, empathy assessment, emotional analysis, moral value recognition, connotation frame analysis, and subjectivity evaluation~\cite{Volkova2018, Rashkin2017, Garten2016, Lee2022, Devlin2018}.
By providing a novel content inoculation framework and evaluation analytics, BRIES enables organizations to understand emergent changes in the information environment that could not otherwise be captured in finite time horizons. 

Moreover, as LLMs become increasingly sophisticated tools for both attack generation and detection, their inherent biases and limitations raise important social implications for information ecosystem resilience \cite{Bender2021, Weidinger2022, NATOStratCom2021, NATODynamicResilience2025}. The socio-emotional-cognitive signatures we extract—measuring dimensions such as toxicity, sentiment, empathy, and moral foundations—provide critical insights into how these signatures manifest when LLMs generate adversarial content or attempt to detect manipulation techniques \cite{Kozyreva2024,SemEval2023}. For instance, LLMs demonstrate varying effectiveness in detecting emotionally charged attacks, or may themselves generate attacks that disproportionately leverage particular emotional triggers or moral values. By analyzing these signature patterns across persuasion attack types, detection models, and content domains, BRIES not only quantifies  performance but also addresses the broader ethical considerations of deploying AI systems in our information ecosystem \cite{Zavolokina2024,NATOInfoThreat2024}. 

BRIES delivers quantifiable evidence of defense effectiveness against targeted attacks and offers actionable recommendations to strengthen training materials and organizational resilience against information manipulation. This innovative approach represents a shift from reactive to proactive defense, creating "reputational armor" for individuals and organizations through automated techniques that build and test cognitive resiliency.

\section{Methodology}
The BRIES architecture builds on McGuire's (1964) foundational inoculation theory by implementing its key components: threat awareness and refutational preemption. Our Defender agent extends McGuire's approach from "cultural truisms" to complex digital persuasion techniques, while our Assessor component operationalizes "resistance conferral" through measurable socio-emotional-cognitive signatures. Additionally, BRIES integrates dual-process theories of persuasion~\cite{PettyCacioppo1986,Chaiken1980} by addressing both systematic processing through explicit argumentation and heuristic processing through affective cues. This allows our framework to target distinct vulnerability pathways exploited by different attack types, providing comprehensive protection against modern persuasion techniques that operate through both rational and automatic cognitive processes.

\subsection{Detector}
The Detector agent employs a system of structured prompts to identify persuasive attacks in text-based content. Our framework supports both high-precision detection focused on the most prevalent types of manipulation as well as comprehensive coverage of a broader range of attack techniques. The base detector prompt instructs the language agent to identify specific attack types from a provided taxonomy, following a structured format: {\it ``Identify any logical fallacies in the following text. Here is a list of common fallacies to consider: [fallacies]. Text: [attack text].''}

To enhance detection capabilities, we developed several prompt variants. The {\it descriptions-enhanced prompt} $d_0$ augments each attack with a detailed explanation, improving the agent's contextual understanding of various attack patterns. For {\it confidence assessment} $s_0$, we implemented a scoring prompt requiring the agent to rate its confidence in each identified attack on a scale of 1 to 10, resulting in outputs following the format {\it "attack - score: \#"} where higher scores indicate greater confidence. Our most sophisticated prompt combines attack descriptions with confidence scoring, merging contextual benefits with quantitative assessment. The technical implementation of the detector integrates with modern LLM infrastructure through the OpenAI API, using Ollama for local model hosting and LiteLLM Proxy Server to facilitate seamless routing between agents. 

\subsection{Defender}
The Defender agent builds upon the outputs of the Detector to generate content that is resistant to persuasive attacks. Our initial implementation incorporated both the original content and attacks identified by the detector, leveraging prompt engineering to generate inoculated content that preserves the  integrity of the source  while strengthening it against potential manipulation: {\it Read the original text and the list of fallacies found within the text and then rewrite the text to correct the issues. Make sure the rewritten text retains all of the same information as the original text. Only respond with the rewritten text. Do not add explanations or formatting.
        Original Text: {article["attack"]}
        Attack: {detector response}}.

The Defender agent implements techniques from inoculation theory \citep{McGuire1961, Pfau1997}, which posits that exposing individuals to weakened forms of persuasive attacks helps build resistance to stronger future attempts. By incorporating both the threat component (awareness of vulnerability) and refutational preemption (explicit countering of potential arguments), our defender creates content that not only withstands specific known attacks but also builds general resilience against novel manipulation techniques \citep{Compton2021}.

\subsection{Assessor}
The Assessor tools come at the end of our compound AI workflow for content inoculations, after the Detector and Defender agents. By analyzing inputs and outputs from these agents, the Assessor analyzes the effect BRIES agents have on target outcomes e.g., the effectiveness of inoculations from the defender, given the characteristics of the Defender. To do so, we leverage causal machine learning (ML) tools to (1) explore causal linkages and structure (e.g., “when $\boldsymbol{X}$ is seen to increase, we see a decrease in $\boldsymbol{Y}$”) and (2) estimate the average treatment effect (ATE) seen in a target metric given some intervention. Causal analytics allow the Assessor to move beyond correlation-based models by relying on those built around Pearl’s \citep{pearl2009causality, pearl2018bookofwhy} causal framework. Doing so makes the Assessor proactive, rather than reactive, in determining possibilities of behavior seen around media that has been manipulated to foster persuasion or influence the reader. Below we briefly detail the methods used for BRIES’s Assessor tools.
\begin{itemize}
  \item \textbf{Structural Equation Modeling (SEM)}: The tool relies on the NOTEARS algorithm and the package Causalnex \citep{quantumblack2020causalnex, zheng2018notears} to perform causal discovery in the form of directed-acyclic graph (DAG) weights and edges. We focus on treatment and outcome's (e.g., socio-emotional-cognitive and detection metrics) currently, blocking incoming edges from our treatments for logical reasons. Thus the current experiments  avoid confounding effects where variables that effect outcome also affect treatments at the same time.
  \item \textbf{Average Treatment Effect Estimation (ATE)} During this study we focused on ATE estimation via causal forests from the package EconML \citep{battocchi2019econml, chernozhukov2016double, wager2018estimation}, where the input details included treatments, outcomes, and co-variates (e.g., measures from our agent pipeline outside the set of treatments and Defender agent measures). During ATE estimation we remove all but one treatment at a time to simplify the analysis and reduce computational burdens. 
\end{itemize}

\section{Attack Detection Results}
Across all models and prompting strategies, clear patterns emerged regarding which attack types were consistently well-detected versus those that proved challenging:
\begin{itemize}
    \item {\it Easily Detected Attacks:} Appeal to Authority, Appeal to Fear, and Flag Waving consistently achieved high F1 across models.
    \item {\it Moderately Detected Attacks:} Appeal to Values, Appeal to Popularity, and False Dilemma showed moderate detectability with significant variation across models.
    \item {\it Challenging Attacks:} Red Herring, Conversation Killer, and Questioning the Reputation proved extremely difficult to detect reliably, often scoring below $F1<0.20$ across multiple models.
\end{itemize}

 \begin{figure}[b!]
 \centering
  \includegraphics[width=0.5\textwidth]{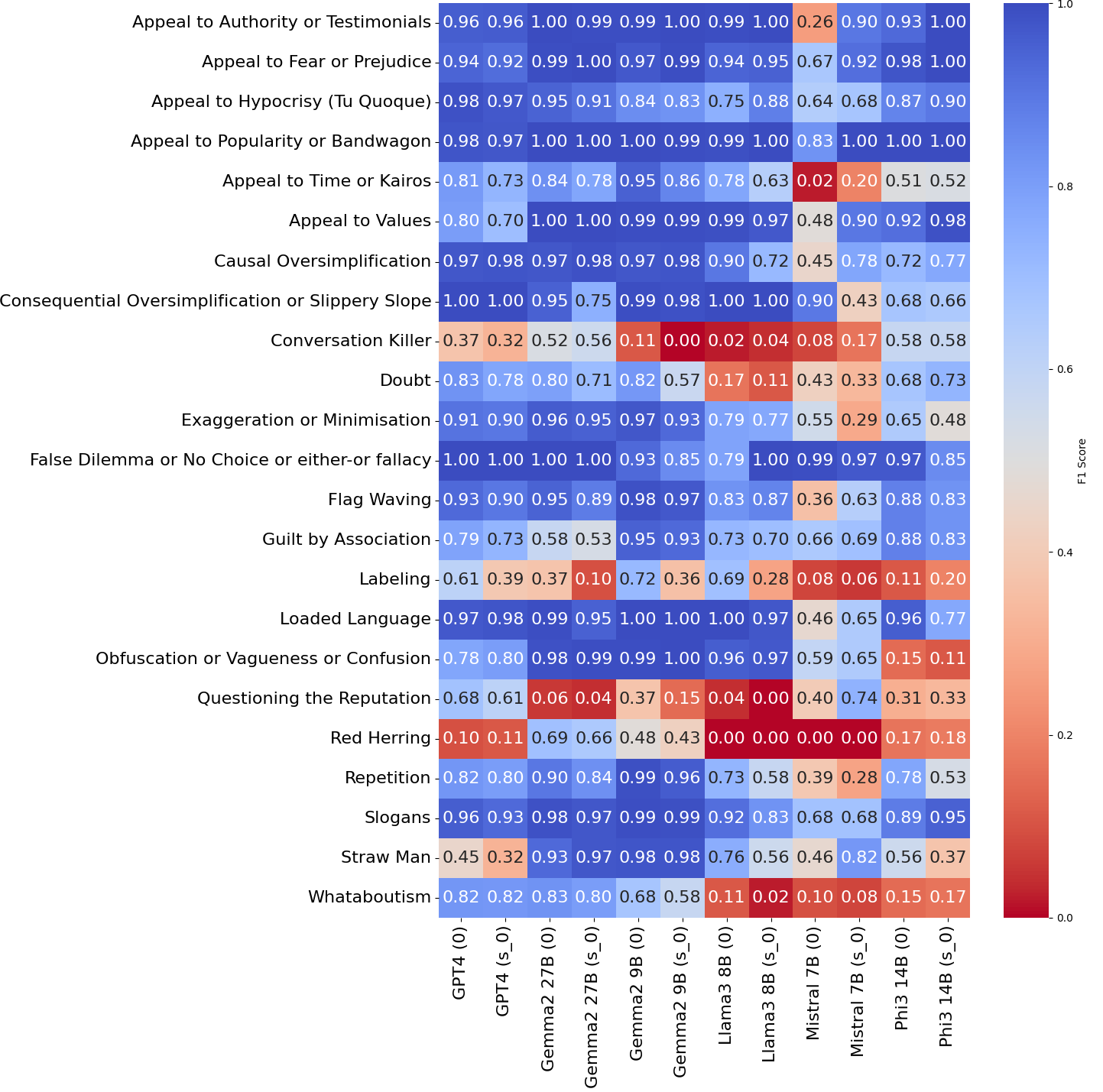}
 \caption{Persuasion attack detection results across 23 attack types for open (Gemma2, LLaMa3, Mistral and Phi3) vs. closed models (GPT4) with ($s_0$) = asking for a confidence score, (0) not asking.}
 \label{fig:confidence}
 \end{figure}

\subsection{ Impact of Confidence Score Requests} Requesting confidence scores produced distinctly different effects across model architectures (Figure ~\ref{fig:confidence}). For Gemma2 27B, strategy $s_0$ (not requesting confidence scores) generally performed better, particularly in detecting Labeling ($0.37$ vs $0.10$) and Questioning the Reputation ($0.06$ vs $0.04$). GPT-4 showed a clear preference for the strategy with consistently higher scores without confidence requests, especially for Appeal to Values ($0.80$ vs$ 0.70$) and Straw Man arguments ($0.45$ vs $0.32$). Interestingly, Llama3 8B demonstrated improved performance with strategy $s_0$ (requesting confidence scores), notably for False Dilemma ($0.79$ vs $1.00$) and Appeal to Authority ($0.99$ vs $1.00$). Mistral 7B showed the most improvement with $s_0$ strategy, particularly with complex persuasion techniques like Appeal to Authority ($0.26$ vs $0.90$) and Flag Waving ($0.36$ vs $0.63$).

\subsection{Effect of Attack Descriptions in Prompts}
\begin{figure}[t!]
 \centering
  \includegraphics[width=0.5\textwidth]{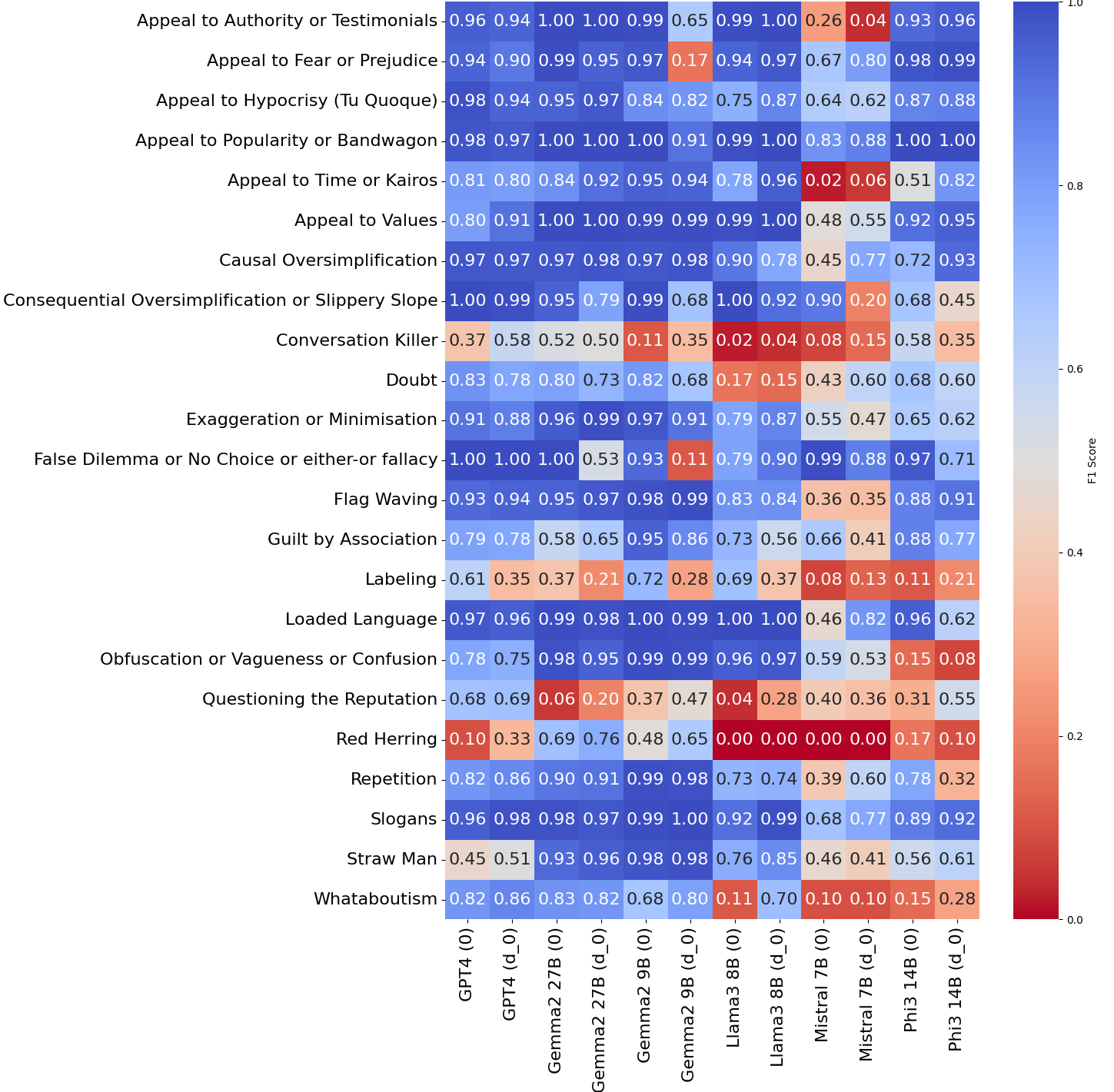}
 \caption{Persuasion attack detection results across 23 attack types for open vs. closed models with (0) models use the base prompt while ($d_0$) use a modified prompt that includes a description of each attack listed. Models have temperature set to 0.}
 \label{fig:description}
 \end{figure}

The analysis of base (0) versus descriptive ($d_0$) prompting revealed model-specific optimal strategies (Figure ~\ref{fig:description}). For Gemma2 27B, the base prompt (0) consistently outperformed the descriptive prompt ($d_0$), particularly evident in techniques like False Dilemma ($1.00$ vs $0.53$) and Labeling ($0.37$ vs $0.21$). GPT-4 showed relatively stable performance across both strategies but with a slight advantage using the base prompt (0). The Llama 3 8B model demonstrated better performance with the descriptive prompt ($d_0$), showing improved detection in categories like Whataboutism ($0.11$ vs $0.70$). Similarly, the Mistral 7B model generally yielded better results with the descriptive prompt ($d_0$), particularly for techniques like Appeal to Authority ($0.26$ vs $0.04$). The Phi3 14B model showed mixed results but generally performed better with the base prompt, especially for techniques like Obfuscation ($0.15$ vs $0.08$).

\subsection{Temperature Effects on Performance} 
\begin{figure}[t!]
 \centering
  \includegraphics[width=0.5\textwidth]{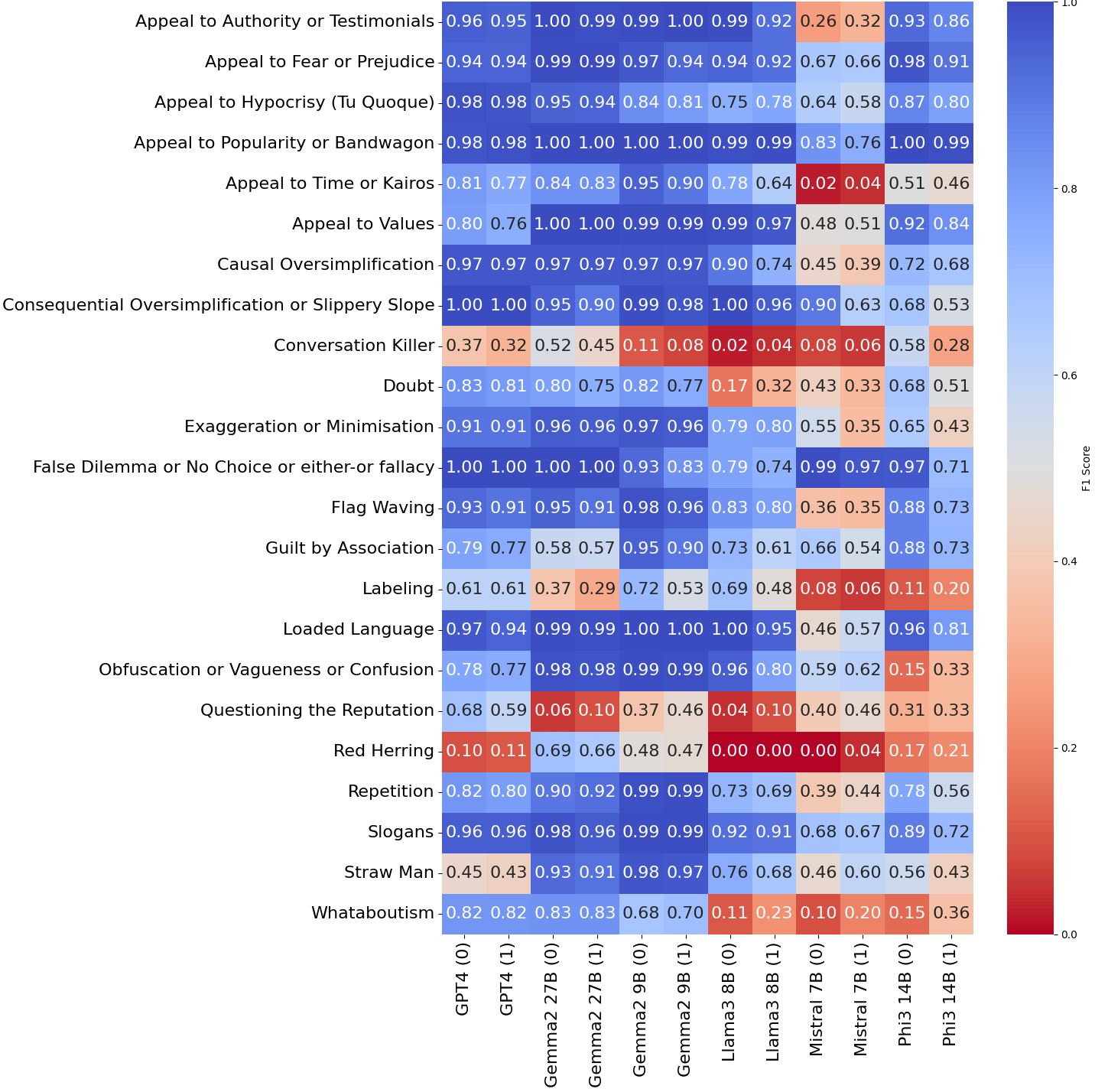}
 \caption{Persuasion attack detection results across 23 attack types for open vs. closed models (GPT4) with temperatures set up to (0) or (1).}
 \label{fig:temperature}
 \end{figure}

Our analysis revealed significant model-specific responses to temperature settings (Figure~\ref{fig:temperature}). The Gemma family (Gemma2 27B and Gemma2 9B) demonstrated marginally better performance at temperature $0$, indicating inherent stability across temperature variations. GPT-4 showed a clear preference for temperature $0$, maintaining exceptional F1 scores consistently above $0.90$ across most persuasion attacks. In contrast, the Llama3 8B and Mistral 7B  models performed signiticantly better at temperature $1$, with improved detection of nuanced persuasion techniques such as Doubt and Whataboutism. The Phi3 14B model aligned with the GPT-4 and Gemma patterns, showing optimal performance at temperature 0 with higher F1 scores across most categories. Overall, our analysis suggests that larger, more sophisticated models (GPT-4) benefit from simpler prompts, while smaller or differently architected models (Llama, Mistral) perform better with additional context and explicit confidence requests.

\section{Inoculation Effectiveness Analysis Results}
Our analysis of content inoculation effectiveness demonstrates how various factors influence the resilience of content against persuasive attacks. Specifically, we examine the effect of Detector and Defender LLMs and persuasion attacks on socio-emotional-cognitive (SEC) measures. Note, for the causal investigations we extended the set of LLMs beyond those used in our detection experiments.

\subsection{LLM Effects on Content Inoculations}
SEM assessor tools reveal consistent patterns across LLM interventions create resilient content (Table~\ref{tab:llm_effects_horizontal}). All examined LLMs demonstrate strong positive effects on emotional dimensions, particularly sadness (values $~0.76$) and neutral perspective (values $~0.72$0. These models simultaneously show moderate positive effects on positive perspective (values $~0.3$) while negatively impacting attitude measures (values $-0.3$ to $-0.4$). GPT-family models, particularly GPT-4o ($0.48$), demonstrate substantially stronger positive effects on detection compared to open-source alternatives, suggesting their superior capability for identifying manipulative content.

\begin{table*}[t!]
  \centering
  \caption{Significant LLM Effects on Content Inoculation Outcomes ($|value| > 0.05$)}
  \resizebox{\textwidth}{!}{%
  \begin{tabular}{l|rr|rr|rr|rr|rr|rr|rr|rr}
    \toprule
    \textbf{SEC Outcomes} & \multicolumn{2}{c|}{\textbf{GPT4o}} & \multicolumn{2}{c|}{\textbf{GPT4o Preview}} & \multicolumn{2}{c|}{\textbf{GPT4 Turbo}} & \multicolumn{2}{c|}{\textbf{GPT3.5 Turbo}} & \multicolumn{2}{c|}{\textbf{Llama3 8B}} & \multicolumn{2}{c|}{\textbf{Mistral 7B}} & \multicolumn{2}{c|}{\textbf{Phi3 4B}} & \multicolumn{2}{c}{\textbf{Gemma 7B}} \\
    & \textbf{ATE} & \textbf{SEM} & \textbf{ATE} & \textbf{SEM} & \textbf{ATE} & \textbf{SEM} & \textbf{ATE} & \textbf{SEM} & \textbf{ATE} & \textbf{SEM} & \textbf{ATE} & \textbf{SEM} & \textbf{ATE} & \textbf{SEM} & \textbf{ATE} & \textbf{SEM} \\
    \midrule
    Attitude & & 0.387 & 0.211 & 0.559 & & 0.453 & & 0.434 & -0.074 & 0.457 & -0.212 & 0.354 & 0.172 & 0.501 & -0.123 & 0.401 \\
    Emotion: Sadness & & 0.759 & & 0.759 & & 0.759 & & 0.759 & & 0.759 & & 0.759 & & 0.760 & & 0.759 \\
    Intent: Agreeing & -0.069 & 0.066 & & 0.130 & & 0.145 & -0.084 & 0.070 & 0.059 & 0.175 & -0.057 & 0.107 & 0.070 & 0.196 & 0.078 & 0.176 \\
    Moral: Authority & & 0.104 & & 0.122 & & 0.147 & & 0.117 & & 0.108 & 0.067 & 0.173 & & 0.112 & & 0.154 \\
    Moral: Fairness & & 0.052 & & 0.054 & & & & 0.053 & & 0.074 & & 0.054 & & & & \\
    Moral: Harm & & 0.296 & & 0.312 & & 0.297 & & 0.300 & & 0.327 & & 0.293 & & 0.256 & & 0.276 \\
    Moral: Ingroup & & 0.094 & & 0.061 & & 0.078 & & 0.067 & -0.053 & & & 0.072 & 0.097 & 0.116 & & 0.074 \\
    Morality & & & & & & & & 0.063 & & 0.078 & & 0.070 & & 0.067 & & 0.059 \\
    Perspective: Neutral & & 0.718 & & 0.718 & & 0.718 & & 0.717 & & 0.718 & & 0.718 & & 0.718 & & 0.718 \\
    Perspective: Positive & & 0.317 & & 0.301 & & 0.300 & & 0.310 & & 0.290 & & 0.281 & & 0.290 & & 0.302 \\
    Sentiment & & & 0.103 & & & & & & -0.062 & & -0.098 & & 0.082 & & -0.140 & \\
    \bottomrule
  \end{tabular}
  }
  \label{tab:llm_effects_horizontal}
\end{table*}

When assessed through ATE estimation, LLMs show varying effectiveness. GPT-4o and GPT-4 Turbo demonstrate strong positive effects on detection ($+0.24$ and $+0.19$ respectively), while others like GPT-3.5 Turbo show significant negative effects ($-0.24$). Models also exhibit distinct patterns in their effects on moral and emotional dimensions, with GPT-4o Preview showing positive effects on Attitude ($+0.21$) and Sentiment ($+0.10$), suggesting enhanced ability to maintain positive framing while improving resilience.

\subsection{Persuasion Attack Effects on Content Inoculations}
Our analysis of persuasion attack techniques reveals more diverse and targeted effects (Table~\ref{tab:attack_effects_horizontal_full}). Appeal to Authority demonstrates an exceptionally strong positive effect on Detection ($0.52$), while distraction-based techniques like Red Herring show pronounced negative effects ($-0.29$). A clear pattern emerges where techniques designed to distract or obfuscate consistently reduce detection capabilities, while those appealing to emotions or presenting false choices increase awareness. All attack interventions show consistent positive effects on Sadness (values $~0.26$) and Neutral perspective (values $~0.25$), suggesting these emotional shifts are fundamental outcomes of rhetorical interventions.

The ATE analysis confirms these findings, with Appeal to Authority showing the strongest positive effect on detection ($+0.43$), followed by Appeal to Fear ($+0.30$). Conversely, techniques like Red Herring ($-0.39$), Labeling ($-0.36$), and Questioning the Reputation ($-0.31$) show strong negative effects on detection, highlighting specific vulnerabilities that require specialized inoculation approaches.

\begin{table*}[htbp]
  \centering
  \caption{Significant Persuasion Attack Effects on Outcomes ($|value| > 0.05$)}
  \small
  \resizebox{\textwidth}{!}{%
  \begin{tabular}{l|rr|rr|rr|rr|rr|rr|rr|rr|rr}
    \toprule
    \textbf{Attack/Intervention} & \multicolumn{2}{c|}{\textbf{Attitude}} & \multicolumn{2}{c|}{\textbf{Intent:}} & \multicolumn{2}{c|}{\textbf{Moral:}} & \multicolumn{2}{c|}{\textbf{Moral:}} & \multicolumn{2}{c|}{\textbf{Moral:}} & \multicolumn{2}{c|}{\textbf{Moral:}} & \multicolumn{2}{c|}{\textbf{Morality}} & \multicolumn{2}{c|}{\textbf{Perspective:}} & \multicolumn{2}{c}{\textbf{Sentiment}} \\
    & & & \multicolumn{2}{c|}{\textbf{Agreeing}} & \multicolumn{2}{c|}{\textbf{Authority+}} & \multicolumn{2}{c|}{\textbf{Fairness+}} & \multicolumn{2}{c|}{\textbf{Harm+}} & \multicolumn{2}{c|}{\textbf{Ingroup+}} & & & \multicolumn{2}{c|}{\textbf{Positive}} & \\
    & \textbf{ATE} & \textbf{SEM} & \textbf{ATE} & \textbf{SEM} & \textbf{ATE} & \textbf{SEM} & \textbf{ATE} & \textbf{SEM} & \textbf{ATE} & \textbf{SEM} & \textbf{ATE} & \textbf{SEM} & \textbf{ATE} & \textbf{SEM} & \textbf{ATE} & \textbf{SEM} & \textbf{ATE} & \textbf{SEM} \\
    \midrule
    Appeal to Authority & -0.183 & -0.133 & & 0.108 & & & & & & 0.094 & & & & & & 0.087 & & \\
    Appeal to Fear & 0.154 & 0.463 & -0.055 & & & 0.051 & & & 0.050 & 0.162 & & & & & & 0.127 & 0.109 & \\
    Appeal to Hypocrisy & 0.069 & 0.205 & & 0.130 & & & & & & & 0.067 & 0.080 & & & & 0.104 & & \\
    Appeal to Time & -0.068 & 0.101 & 0.057 & 0.051 & & & & & & 0.132 & & & & & & 0.112 & & \\
    Appeal to Values & 0.109 & 0.064 & & 0.087 & & 0.051 & & 0.077 & & 0.117 & & & & 0.113 & & 0.100 & 0.093 & \\
    Bandwagon & & 0.117 & 0.098 & & 0.079 & 0.123 & & & & 0.117 & & & & & & 0.098 & 0.067 & \\
    Causal Oversimplification & & 0.147 & -0.177 & -0.120 & & 0.073 & & & & 0.058 & & & & & & 0.108 & & \\
    Conversation Killer & -0.056 & -0.081 & & 0.135 & & 0.090 & & & 0.065 & 0.079 & 0.124 & 0.069 & & & & 0.114 & & \\
    Doubt & & 0.178 & & 0.226 & & & & & & 0.081 & & & & & & 0.090 & & \\
    Exaggeration or Minimisation & & 0.067 & & -0.051 & & & & & & 0.064 & & 0.069 & & & & 0.097 & & \\
    False Dilemma & 0.373 & 0.702 & -0.066 & & & -0.054 & & & & 0.136 & & & & & & 0.091 & & \\
    Flag Waving & -0.208 & & 0.098 & & & 0.097 & & & & 0.109 & & 0.056 & & & & 0.100 & & \\
    Guilt by Association & 0.124 & 0.409 & -0.057 & 0.069 & & & 0.060 & & & 0.059 & & 0.082 & & & & 0.088 & & \\
    Labeling & -0.073 & 0.078 & & & 0.062 & 0.080 & & & -0.072 & & & & & & & 0.115 & -0.067 & \\
    Loaded Language & -0.161 & & & & & & & & & 0.140 & & & & & & 0.123 & & \\
    Obfuscation & 0.058 & 0.204 & & & & 0.081 & & & 0.087 & 0.161 & & -0.083 & -0.053 & & & 0.114 & & \\
    Questioning the Reputation & -0.156 & -0.220 & 0.093 & 0.188 & 0.054 & 0.052 & & & & 0.114 & & & & 0.086 & & 0.077 & & \\
    Red Herring & & 0.236 & & & & 0.079 & & & & 0.064 & & & & & & 0.086 & & \\
    Repetition & -0.088 & 0.087 & -0.067 & -0.172 & & 0.054 & & & & 0.089 & & 0.071 & & & & 0.102 & & \\
    Slippery Slope & 0.135 & 0.377 & & 0.088 & & & & & 0.101 & 0.180 & -0.082 & & & & & 0.119 & 0.119 & \\
    Slogans & 0.062 & 0.195 & -0.103 & -0.222 & & & & & & 0.053 & 0.060 & 0.080 & & & & 0.129 & & \\
    Straw Man & & -0.111 & 0.098 & 0.297 & & & & & & 0.113 & & & & & & 0.115 & & \\
    Whataboutism & 0.182 & 0.352 & & 0.181 & & 0.059 & & 0.050 & & 0.143 & & & & & & 0.093 & & \\
    \bottomrule
  \end{tabular}
  }
  \label{tab:attack_effects_horizontal_full}
\end{table*}

\subsection{Moral Foundation and Intent Outcomes}
Our analysis reveals significant moral foundation effects across both LLM and attack interventions. Among attack techniques, Slippery Slope demonstrates a strong positive effect on Moral: Harm+ ($+0.10$), while Guilt by Association strongly affects Moral: Fairness+ $(+0.06$). Techniques like Flag Waving and Questioning the Reputation show strong positive effects on Intent: Agreeing ($+0.10$ and $+0.09$, respectively), suggesting they increase compliance with messaging — a key vulnerability pathway that inoculation strategies must address.

\section{Discussion}
By understanding  complex relationships between Defender and Detector LLMs, persuasion attacks, and their effects on socio-emotional-cognitive dimensions, we can design more effective content inoculation strategies that target specific vulnerability pathways and strengthen cognitive defenses against information campaigns.

\paragraph{Causal Tool Complementarity}
The comparative analysis of SEM and ATE methodologies reveals complementary insights into content inoculation dynamics. SEM excels at capturing complex interrelationships and cascading effects across socio-emotional-cognitive dimensions, demonstrating how persuasion techniques function within a network of psychological responses, with techniques like Appeal to Authority showing amplified effects ($0.52$ in SEM vs. $0.43$ in ATE) when accounting for these relationships. Conversely, ATE provides more precise isolation of direct causal impacts in controlled contexts, offering clearer measurement of individual intervention effects while controlling for confounding variables. This methodological complementarity is particularly evident in detecting technique-specific pathways—SEM revealed consistent secondary effects on emotional states (Sadness $~0.26$) across all techniques that might be overlooked in ATE analysis, while ATE identified stronger negative effects of distraction techniques like Red Herring ($-0.39$) than captured in SEM. Together, these approaches provide a comprehensive framework for understanding both the isolated efficacy of specific inoculation strategies and how these strategies function within the complex psychological ecosystem of persuasion resistance, enabling more sophisticated, targeted approaches to building cognitive resilience against information manipulation \citep{pearl2009causality, battocchi2019econml, wager2018estimation}.

\subsection{Implications for Content Inoculation Strategies}
\paragraph{Context-Dependent Effects}
Our results demonstrate that persuasion techniques function differently in isolation versus within a broader context of relationships. This suggests that inoculation strategies must account for how persuasion techniques interact with other factors rather than treating each technique as independent. This is  evident in the way techniques like Whataboutism show modest effects on individual measures in ATE analysis ($0.18$ on Attitude) but reveal stronger networked effects when analyzed through SEM ($0.35$ on Attitude), suggesting their impact is amplified through cascading psychological pathways. 
Detection models should therefore be trained on complex, multi-technique content rather than simplified examples of individual techniques. Implementation should combine the strengths of different LLM architectures—utilizing GPT models for techniques they excel at detecting (Appeal to Authority, False Dilemma) while employing open-source models like Mistral for their enhanced sensitivity to emotional manipulation techniques.

\paragraph{Model-Specific Detection Considerations}
Different LLMs show varying patterns of effectiveness at detecting specific persuasion techniques, suggesting that Detection agent should be customized based on which techniques particular models are most vulnerable to. Our data reveals that models like Mistral 7B and Llama3 8B demonstrate higher sensitivity to intent and morality dimensions in SEM analysis ($-0.06$ and $0.06$ on Intent: Agreeing respectively), suggesting they may require specific inoculation against manipulation targeting intent pathways. GPT-family models show stronger detection capabilities overall but exhibit specific vulnerabilities. For instance, GPT-4o Preview shows significant effects on Attitude ($0.21$ in ATE vs. $0.56$ in SEM), indicating a potential overreliance on attitudinal measures that could be exploited by sophisticated persuasion attacks. These findings suggest that optimal detection agents should implement model ensembles that compensate for individual model weaknesses while capitalizing on their respective strengths.

\paragraph{Technique-Specific Inoculation Strategies}
Our comparative analysis identifies specific persuasion techniques requiring targeted defensive approaches.
\begin{itemize}
\item {\bf Authority-Based Techniques} Content should be inoculated against Appeal to Authority and Appeal to Popularity by emphasizing critical evaluation of source credibility. Since these techniques show higher impact in SEM analysis (Appeal to Authority: $-0.13$ on Attitude in SEM vs. $-0.18$ in ATE), contextual cues about authority should be  addressed in inoculation training.

\item {\bf Emotional Manipulation} Defenses against techniques like Guilt by Association and Appeal to Fear should focus on emotional awareness training. The significant ATE impact of Appeal to Fear on Sentiment ($0.11$) and Moral: Harm+ ($0.05$), compared to its absence in SEM suggests that direct emotional effects may precede their integration into broader psychological networks.

\item {\bf Distraction Techniques} Red Herring's unexpected positive effect on Attitude in SEM ($0.24$) despite showing no significant ATE suggests this technique operates through indirect pathways, potentially by distracting from negative aspects of content. Inoculation strategies should therefore include training to recognize subtle distractions that appear positive but undermine critical evaluation.

\item {\bf Agreement Manipulation} Straw Man's consistent effectiveness across both methodologies ($0.30$ on Intent: Agreeing in SEM, $0.10$ in ATE) suggests a need for robust reasoning models that specifically address misrepresentation tactics. Similarly, Slogans' enhanced persuasive power in contextual settings indicates a need for critical language analysis in inoculation training.
\end{itemize}

\subsection{Practical Applications for Building Resilience}
Our findings translate directly into practical applications for building resilient information ecosystems:
\begin{itemize}
\item {\bf Tailored Inoculation Training Programs} Organizations should develop training programs that address the specific vulnerability pathways identified in our analysis. For example, training for contexts where Authority-based manipulation is likely should emphasize source evaluation techniques, while contexts vulnerable to emotional manipulation should incorporate emotional awareness.

\item {\bf LLM-Specific Deployment Strategies} When using LLMs for content evaluation and inoculation, organizations should select models based on their specific strengths. GPT-family models excel at detecting direct persuasive attacks but may miss more subtle emotional manipulations that open-source models like Mistral 7B detect more effectively.

\item {\bf Comprehensive Detection Frameworks} Effective detection systems should combine both SEM and ATE methodologies, using ATE for initial screening of direct effects and SEM for understanding how these effects propagate through psychological networks. This dual approach provides a more complete understanding of vulnerability pathways.

\item {\bf Dynamic Inoculation Content} Rather than static training materials, organizations should develop dynamic content that adapts to specific attack vectors and vulnerability patterns. For instance, content designed to build resilience against Appeal to Fear should address both direct emotional impacts (identified through ATE) and secondary effects on moral reasoning (revealed via SEM).
\end{itemize}

By implementing these practical applications, organizations can significantly enhance their resilience against sophisticated persuasion attacks, protecting both individuals and institutional integrity within increasingly complex information ecosystems \citep{Compton2021, VanderLinden2017}.

\paragraph {Ethical Considerations} While BRIES advances cognitive security capabilities, it presents several ethical challenges. First, our approach faces inherent tension between protecting information integrity and respecting diverse perspectives, risking potential suppression of legitimate dissent if deployed without appropriate safeguards. Second, the use of causal inference on socio-emotional-cognitive signatures raises privacy concerns that must be balanced against collective security benefits. Finally, we acknowledge the dual-use nature of our research, as techniques developed for defense could potentially be repurposed for more sophisticated attacks, necessitating careful consideration of deployment contexts and governance frameworks.

\section{Related Work}
Inoculation theory, originally developed by McGuire \cite{McGuire1964}, has evolved from focusing on cultural truisms to addressing contemporary challenges like information manipulations \cite{VanderLinden2017} and contested science \cite{Compton2021}. However, existing applications have primarily used static content rather than adaptive, AI-generated inoculations. For persuasion attack detection, recent benchmarks like SemEval 2023 Task 3 \cite{SemEval2023} have established taxonomies for evaluating systems, though most operate as standalone detectors rather than integrated defense frameworks. Agent-based social simulations such as SOTOPIA \cite{Zhou2024SOTOPIA} demonstrate LLMs' capabilities for complex interactions, but aren't specifically designed for persuasion resilience. Military approaches to cognitive security and resilience~\cite{fitzpatrick2022information}, including NATO's Cognitive Warfare Concept \cite{NATOACT2023,NATOStratCom2021}, provide strategic frameworks but often lack scalable technical implementations. Meanwhile, causal inference methodologies \cite{pearl2009causality, wager2018estimation} have advanced intervention assessment but are rarely applied to measuring cognitive resilience.

BRIES advances the state of the art through: (1) an integrated multi-agent architecture connecting detection, defense, and assessment in a unified workflow; (2) systematic comparison of multiple LLM families across varied prompting strategies; (3) application of causal inference methodologies (SEM and ATE) to measure intervention effectiveness; and (4) multi-dimensional socio-emotional-cognitive signature analysis that enables targeted inoculation strategies. BRIES compound AI approach creates synergies between components that previous standalone systems could not achieve, resulting in more effective and adaptive protection against persuasive attacks.

\section{Summary}
The BRIES solution represents a significant advancement in cognitive security and resilience by integrating detection, defense, and assessment agents, models and tools through a compound AI architecture. Our research reveals critical insights into model-specific vulnerabilities to persuasion attacks, with GPT-4 demonstrating superior detection of complex techniques ($F1 > 0.90$ for Appeal to Fear) while open-source models struggle with subtle manipulations ($F1 < 0.30$ for Equivocation). Further, our causal analysis of socio-emotional-cognitive signatures provides understanding of how different attack types target specific cognitive dimensions through distinct pathways, enabling more precise inoculation strategies. By quantifying these vulnerabilities and developing targeted interventions grounded in psychological inoculation theory, BRIES offers a promising approach that harnesses AI not just for threat detection but for actively building human cognitive resilience against persuasive manipulation in an increasingly complex information environment. 

Future work will focus on extending the BRIES architecture to incorporate agent-based population simulations. We plan to develop and evaluate scalable inoculation content delivery mechanisms that can reach diverse populations through multiple channels while adapting to individual cognitive profiles and vulnerability patterns.

\section{Acknowledgments}
This research was developed with funding from the Defense Advanced Research Projects Agency (DARPA). The views, opinions and/or findings expressed are those of the author and should not be interpreted as representing the official views or policies of the Department of Defense or the U.S. Government. Distribution Statement "A" (Approved for Public Release, Distribution Unlimited).

\bibliography{mybibfile}

\end{document}